\documentclass[conference]{IEEEtran}
\IEEEoverridecommandlockouts
\usepackage{cite}
\usepackage{amsmath,amssymb,amsfonts}
\usepackage{algorithmic}
\usepackage{graphicx}
\usepackage{textcomp}
\usepackage{xcolor}
\usepackage{booktabs}
\usepackage{multirow}
\usepackage{algorithm}
\usepackage{pgfplots}
\usepackage{pgfplotstable}
\usepackage{tabularx}
\def\BibTeX{{\rm B\kern-.05em{\sc i\kern-.025em b}\kern-.08em
    T\kern-.1667em\lower.7ex\hbox{E}\kern-.125emX}}
\pgfplotsset{compat=1.18}
\begin{document}

\title{STEM Rebalance: A Novel Approach for Tackling Imbalanced Datasets using SMOTE, Edited Nearest Neighbour, and Mixup\\
}

\author{\IEEEauthorblockN{1\textsuperscript{st} Yumnah Hasan}
\IEEEauthorblockA{\textit{University of Limerick}\\
Limerick, Ireland \\
Yumnah.Hasan@ul.ie}
\and
\IEEEauthorblockN{2\textsuperscript{nd} Fatemeh Amerehi}
\IEEEauthorblockA{\textit{University of Limerick}\\
Limerick, Ireland \\
Fatemeh.Amerehi@ul.ie}
\and
\IEEEauthorblockN{3\textsuperscript{rd} Patrick Healy}
\IEEEauthorblockA{\textit{University of Limerick}\\
Limerick, Ireland \\
Patrick.Healy@ul.ie}
\and
\IEEEauthorblockN{3\textsuperscript{rd} Conor Ryan}
\IEEEauthorblockA{\textit{University of Limerick}\\
Limerick, Ireland \\
Conor.Ryan@ul.ie}

}

\maketitle

\begin{abstract}
Imbalanced datasets in medical imaging are characterized by skewed class proportions and scarcity of abnormal cases.  When trained using such data, models tend to assign higher probabilities to normal cases, leading to biased performance.   Common oversampling techniques such as SMOTE rely on local information and can introduce marginalization issues. This paper investigates the potential of using Mixup augmentation that combines two training examples along with their corresponding labels to generate new data points as a generic vicinal distribution.
To this end, we propose  \emph{STEM}, which combines SMOTE-ENN and Mixup at the instance level. This integration enables us to effectively leverage the entire distribution of minority classes, thereby mitigating both between-class and within-class imbalances. We focus on the breast cancer problem, where imbalanced datasets are prevalent. The results demonstrate the effectiveness of  STEM, which achieves AUC values of 0.96 and 0.99 in the Digital Database for Screening Mammography and Wisconsin Breast Cancer (Diagnostics) datasets, respectively.  Moreover, this method shows promising potential when applied with an ensemble of machine learning (ML) classifiers.
\end{abstract}


\section{Introduction}
\label{sec:intro}

An imbalanced dataset refers to classification data where the proportions of different classes are skewed. 
In medical datasets, abnormal cases are typically less common, and data predominantly consist of normal samples resulting in class imbalance problems. When trained using such a dataset, a model tends to assign a higher probability to the normal cases~\cite{li2010learning, xu2023comprehensive}. Synthetic Minority Oversampling Technique (SMOTE)~\cite{chawla2002smote} and its derivatives 
 are widely used as popular oversampling methods to address class imbalance issues~\cite{elreedy2019comprehensive}. However, due to their reliance on local information, these techniques may overlook the overall distribution of the minority class, as they generate minority sample points through random linear interpolation along the line segment connecting minority samples and their neighbors ~\cite{douzas2018effective}. This can lead to distribution marginalization, particularly for the edge points of minority samples~\cite{wang2021research}.

 To deal with these issues and ensure the availability of an evenly distributed generation of sample points,  we employ the
 
    {\footnotesize \textsuperscript{*}979-8-3503-7035-5/23/\$31.00 \textcopyright 2023 IEEE}

 vicinity distribution among minority samples. In general, supervised learning aims to find a function \( f \) that captures the relationship between input data \( x \) and corresponding target values \( y \) from a joint distribution \( P(x, y) \). To accomplish this, a loss \( l \) measures the mismatch between predicted values \( f(x) \) and the actual targets \( y \) across the entire distribution \( P(x, y) \). Minimizing this loss over the distribution is known as Empirical Risk Minimization (ERM). In practice, however, the true distribution \( P(x, y) \) is usually unknown ~\cite{vapnik1999nature}. Therefore typically, the unknown distribution is approximated using the empirical distribution \( P_\text{e}(x, y) \), which is based on the observed dataset consisting of input-output pairs \( (x_i, y_i) \). Each data point contributes to the empirical distribution through a Dirac mass function \( \delta \), assuming the probability masses cluster around specific points. 
 
 Another approach is to estimate the distribution using the vicinity distribution \( P_\text{v}(x, y) \), which replaces the Dirac mass function with a density estimate in the neighborhood of each data point and assumes smoothness around each sample~\cite{chapelle2000vicinal}. By utilizing the vicinity distribution, models are less likely to memorize specific data points and generalize better, improving performance during testing~\cite{vapnik1999nature}. Augmentation is one way to achieve the vicinity distribution, where the original data points are perturbed within their vicinity~\cite{zhang2017mixup, xu2023comprehensive}.  Among the various augmentation techniques, Mixup ~\cite{zhang2017mixup} serves as a valuable data-agnostic data augmentation, acting as a generic vicinal distribution. When sampling from the Mixup vicinal distribution, virtual feature-target vectors are produced as $\tilde{x} = \lambda x_i + (1 - \lambda) x_j$ and $\tilde{y} = \lambda y_i + (1 - \lambda) y_j$, respectively, 
where $(x_i, y_i)$ and $(x_j, y_j)$ are two feature-target vectors randomly drawn from the training data. The hyper-parameter $\lambda \in [0,1]$ follows a Beta distribution ~\cite{mcdonald1995generalization} $\text{Beta} (\alpha, \alpha)$, where $\alpha$ is the hyperparameter that controls the strength of the interpolation; as $\alpha$ approaches 0, it adheres to the ERM principle~\cite{zhang2017mixup}. By incorporating Mixup as a data augmentation technique, the behavior of \(f\) between training examples is more linear. Here behavior refers to the characteristics of \(f\) when it processes different training examples, and how predictions change when the input data change. This linear behavior has the potential to mitigate undesired oscillations when making predictions beyond the scope of the training examples ~\cite{zhang2017mixup}. 

Given the success of Mixup~\cite{ thulasidasan2019mixup} and the reliance of SMOTE ~\cite{chawla2002smote} and its variations on local information, which can lead to a potential oversight of the overall distribution of the minority class~\cite{douzas2018effective}, in this paper, we investigate whether Mixup could address the marginalization issue. To this end, we introduce a hybrid approach called SMOTE  Edited Nearest Neighbour Mixup (STEM), which combines SMOTE-ENN ~\cite{batista2004study} and Mixup~\cite{zhang2017mixup} at the instance level, specifically for abnormal samples. This would enable us to consider the overall distribution of minority classes rather than solely relying on local information from neighborhoods. SMOTE-ENN will be described in the next section. To conduct a thorough comparison, we employ multiple oversampling algorithms to evaluate their efficacy with a focus on the breast cancer problem, where datasets are typically imbalanced using two publicly available datasets, namely, the Digital Database for Screening Mammography (DDSM)~\cite{DDSM} and the Wisconsin-Breast Cancer (Diagnostics) (WBC)~\cite{wolberg1992breast} dataset. 

The rest of the paper is structured as follows.  Section~\ref{sec:literaure} reviews the existing literature on balancing techniques.  Section~\ref{sec:Dataset Preparation} introduces details of the dataset preparation performed in this study. The detailed procedure of the proposed method is described in section~\ref{sec:Methodology}. The experimental setup was carried out to test the performance of the proposed approach described in section~\ref{sec:Experimental details}. The results are discussed in section~\ref{sec:results} using two real medical datasets. Finally, section~\ref{sec:conclude} provides the conclusions of the study.

\begin{figure}[t]
  \centering
  \includegraphics[width=0.43\textwidth]{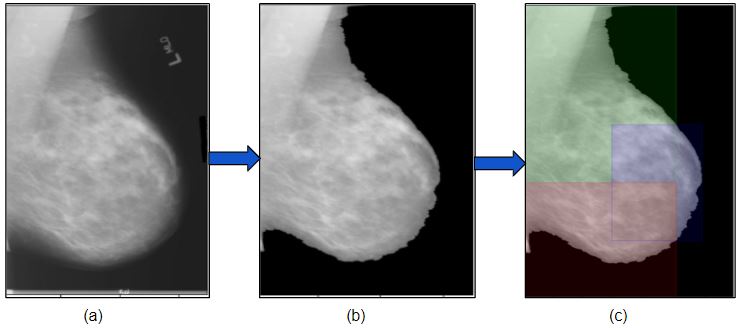} 
  \caption{(a) Original image (b) processed image (c) segmented image.}
     \label{image processing}
\end{figure}

\section{Literature Review}
\label{sec:literaure}
There are three main approaches to tackling the class imbalance problem. The first focuses on modifying or creating algorithms that prioritize learning from the minority class. The second involves applying cost-sensitive techniques at either the data or algorithmic level to minimize errors with higher associated costs. The third revolves around adjusting the data distribution using techniques such as undersampling, oversampling, or combining both to achieve a balanced class distribution~\cite{fernandez2013analysing}.
Undersampling refers to the removal of examples from the majority class, while oversampling involves duplicating or synthesizing examples from the minority class. The combination of the two is referred to as \textit{hybrid}.
%
SMOTE ~\cite{chawla2002smote} can be considered the pioneering oversampling technique, with numerous other variations subsequently developed. In SMOTE, the primary objective is to alleviate the class imbalance by generating synthetic samples for the minority class. This procedure involves identifying the set of minority class samples within the dataset and calculating their $K$-nearest neighbors. 
A random neighbor $\hat{x_i}$ is chosen for each sample $x_i$, and  synthetic sample $x_{\text{new}} = x_i + \delta(\hat{x_i} - x_i)$ is generated through interpolation between the two with a random value $\delta \sim \text{Uniform}[0, 1]$~\cite{chawla2002smote}. Synthetic Minority Over-sampling Technique-Nominal Continuous (SMOTE-NC) is a variation of SMOTE to handle mixed datasets containing both continuous and nominal features.
However, it is not specifically designed to handle datasets that exclusively contain categorical features~\cite{chawla2002smote}.  SMOTE offers the advantage of mitigating the bias in classifiers trained on imbalanced datasets. Nonetheless, SMOTE tends to oversample uninformative and noisy samples~\cite{jiang2021new}.

\begin{table}[t]
\caption{Dataset Description for each setup. S and F indicate segments and full images. Craniocaudal (cc) and mediolateral oblique (mlo) are the two views of mammograms. Tr Pos and Neg are used for training positive and negative samples. CR is the Class Ratio between Pos and Neg instances.}
\label{Table:dataset}
\setlength{\tabcolsep}{10pt}
\begin{tabularx}{\linewidth}{@{}cccccc@{}}
\toprule
\textbf{Dataset}      & \textbf{Setups} & \textbf{Tr Pos} & \textbf{Tr Neg} & \textbf{Total} & \textbf{CR} \\ \midrule
\multirow{4}{*}{DDSM} & $S_{mlo}$             & 89              & 1326            & 1415           & 6:94        \\
                      & $S_{cc}$             & 85              & 1316            & 1401           & 7:93        \\
                      & $S_{cc+mlo}$             & 175             & 2642            & 2817           & 6:94        \\
                      & $F_{cc+mlo}$             & 118             & 822             & 940            & 6:94        \\
WBC                   &                 & 169             & 285             & 454            & 37:63       \\ \bottomrule
\end{tabularx}
\end{table}

Alternative oversampling techniques such as Adaptive Synthetic Sampling method (ADASYN) ~\cite{he2008adasyn} and Borderline-SMOTE~\cite{han2005borderline} may overcome such issues. Borderline-SMOTE prioritizes oversampling and reinforcing minority examples located along class boundaries. It determines whether a minority class instance is eligible for oversampling using the SMOTE by considering if more than half of its $m$ nearest neighbors belong to the majority class.
Consequently, it focuses on enhancing the class distribution by exclusively utilizing minority-class samples located on the boundary to generate new synthetic samples~\cite{han2005borderline}.

Similarly, SVM-SMOTE addresses imbalanced datasets by generating instances specifically along the decision boundary rather than oversampling the entire minority class. To estimate the borderline area, support vectors are derived from training a standard support vector machine (SVM) classifier~\cite{hearst1998support} on the original training set. Then new instances are generated by randomly placing them along the borderline that connects each support vector of the minority class with a set of its nearest neighbors via interpolation or extrapolation based on the density of majority class instances surrounding each support vector ~\cite{nguyen2011borderline}.

A hybrid method, SMOTE-Tomek Links~\cite{zeng2016effective} combines the two techniques of SMOTE and Tomek Links~\cite{tomek1976two}. SMOTE generates synthetic data for the minority class, while Tomek Links identifies and removes data from the majority class that is closely associated with the minority class. 

\begin{algorithm}[ht]
  \caption{Balancing Imbalance using STEM}
  \label{alg:balance_data}
  \begin{algorithmic}[1]
  \renewcommand{\algorithmicrequire}{\textbf{Input:}}
    \renewcommand{\algorithmicensure}{\textbf{Output:}}
    \REQUIRE Imbalance Training Data
    \ENSURE Balanced Training Data
    
    \STATE Apply SMOTE-ENN:
    \STATE \quad Randomly select $x_i$ from minority classes
    \STATE \quad Identify $k$-nearest neighbors of $x_i$ and randomly select one of the neighbors $\hat{x}_i$
    \STATE \quad Generate $x_{\text{new}} = x_i + \delta(\hat{x}_i - x_i)$  where ($\delta \in [0,1]$)
    
    \IF{balancing ratio does not satisfy}
    \STATE Go to step 1
     \ELSE
        \STATE Remove noise samples using ENN:
        \FOR{every instance $x_j$}
            \STATE Find the three nearest neighbors of $x_j$
            \IF{$x_j$ gets misclassified by its three nearest neighbors}
                \STATE Delete $x_j$
            \ENDIF
        \ENDFOR
    \ENDIF
    
    \STATE Apply Mixup:
    \STATE \quad Randomly select $x_i$ and $x_j$ from the same classes
    \STATE \quad Randomly sample $\lambda$ from Beta distribution
    \STATE \quad $\tilde{x} = \lambda x_i + (1 - \lambda) x_j$
    
    \STATE End
\end{algorithmic}
\end{algorithm}

A Tomek link is defined as a pair of neighbors $(x_i, x_j)$ that has the minimal Euclidean distance ($d(.)$), where $x_i$ belongs to the minority class and $x_j$ belongs to the majority class. A pair $(x_i,x_j)$ forms a Tomek link only when there is no sample $x_k$ that satisfies the conditions $d(x_i, x_k) < d(x_i, x_j)$ or $d(x_j, x_k) < d(x_i, x_j)$. SMOTE-Tomek involves selecting random data points from the minority class, calculating their distances to the $k$ nearest neighbors, multiplying the difference with a random number, and adding the result as synthetic samples to the minority class. 

This process is repeated until the desired proportion of the minority class is achieved. If randomly selected data points from the majority class have a nearest neighbor that belongs to the minority class (indicating the presence of a Tomek Link), the link is eliminated~\cite{zeng2016effective}. Similarly, another hybrid method, namely SMOTE-ENN combines the two techniques of SMOTE and Edited Nearest Neighbour (ENN)~\cite{wilson1972asymptotic} methods. ENN is an undersampling technique that removes instances from the majority class. 
SMOTE-ENN initially applies SMOTE to enhance the representation of the minority class by oversampling it within the dataset. Then, ENN identifies and eliminates instances in the augmented dataset that their nearest neighbors misclassify. This joint technique aims to address the class imbalance and effectively remove potentially noisy instances from the majority class ~\cite{batista2004study}. On the other hand, the main concept behind ADASYN involves utilizing weighted distributions for different minority classes of samples, considering their varying levels of difficulty in learning.
This generates a greater number of synthetic data for minority class samples that are more challenging to learn compared to the minority samples that are easier to learn. This way, ADASYN mitigates the learning bias caused by the initial imbalanced data distribution and dynamically adjust the decision boundary to concentrate on the challenging samples that are harder to learn~\cite{he2008adasyn}.

Each of these approaches tackles the issue of between-class imbalance. However, another type of problem is \textit{within-class} imbalance, where sparse or dense subclusters of minority or majority instances exist. Despite the negative impact of both imbalances on standard classifiers' performance, methods for handling the class imbalance problem typically concentrate on correcting the between-class imbalance while neglecting to address the imbalances within each class ~\cite{japkowicz2001concept}. The purpose of this paper is to tackle both types of imbalances simultaneously. To this end, we employ Mixup~\cite{zhang2017mixup} an augmentation technique where two examples from the training dataset are randomly selected, and a new synthetic example is created by linearly combining their feature vectors and labels. Despite its simplicity, Mixup has demonstrated remarkable effectiveness as a data augmentation method. Deep neural networks (DNNs) trained with Mixup have significantly improved classification performance across various image classification benchmarks~\cite{thulasidasan2019mixup}. In this study, we explore the applicability of Mixup at the instance level on balanced data that comes after applying SMOTE-ENN  to ensure that the mixed label remains the same as the original label and yet improves  within-class imbalance. 

\begin{figure}[t]
  \centering
  \includegraphics[width=0.48\textwidth]{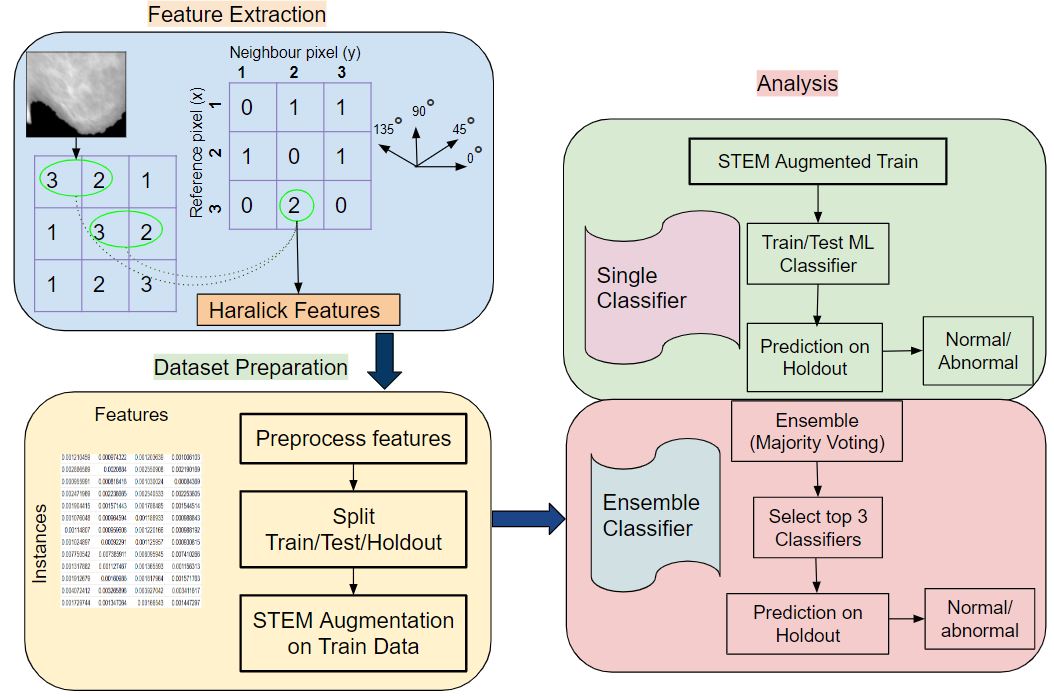} 
  \caption{Proposed approach from feature extraction to image classification using single and ensemble models.}
  \label{methodology}
\end{figure}

\section{Dataset Preparation}
\label{sec:Dataset Preparation}
The proposed algorithm's performance is evaluated using the Digital Database for Screening Mammography (DDSM)~\cite{DDSM} and Wisconsin-Breast Cancer (Diagnostics) (WBC) ~\cite{wolberg1992breast} datasets. Table~\ref{Table:dataset} provides the specifics of each experimental setup's positive and negative segments. 

\begin{figure*}[ht]
    \centering
    \includegraphics[width=0.8\linewidth]{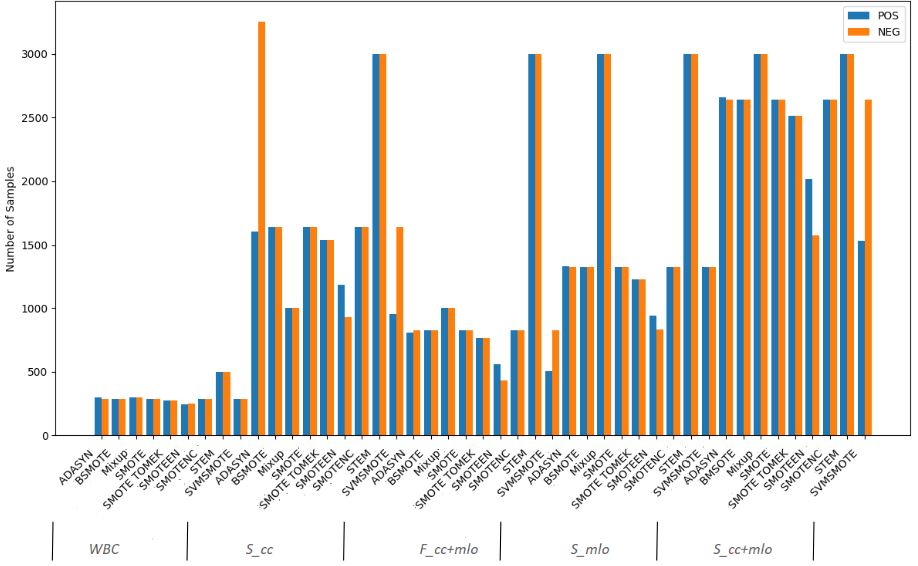}
    \caption{Positive and negative samples for each dataset and augmentation approach}
    \label{sample}
\end{figure*}

\subsection{DDSM}
DDSM is a widely-used public database of digital mammograms containing both normal and abnormal images. It is a significant resource in breast cancer detection and diagnosis, comprising 43 volumes of mammogram samples. These volumes include normal, abnormal, benign, and benign instances without callback cases. All volumes share a consistent patient classification. For this study, we select \textbf{Cancer\_02} and three volumes of normal images \textbf{1-3} while excluding benign or benign without callback cases. To create a realistic domain dataset, we intentionally maintain a high-class imbalance ratio by choosing one volume of cancer images versus three volumes of normal images.

Each case includes images from Craniocaudal (CC) and Mediolateral Oblique (MLO) views for both the left and right sides of the breast. There are 148 positive and 1028 negative images, of which 152 are from the cancer volume, as tumors usually occur on one side of the breast, and 876 are from the normal volumes \textbf{1-3}. After the segmentation step, each image is divided into three segments, namely, top, mid, and bottom, for both views and sides of the breast. This results in four segments per image: the entire breast ($I$), the top segment ($I_t$), the mid-segment ($I_m$), and the bottom segment ($I_b$).

\subsection{WBC}

The WBC dataset is extensively utilized and comprises 30 features extracted from a breast mass's Fine Needle Aspiration (FNA). The patients are classified into benign (non-cancerous) and malignant (cancerous). The dataset contains 569 samples, with 212 labeled as positive, representing malignant cases. Conversely, 357 data points are classified as benign, corresponding to the negative cases.

\section{Methodology}
\label{sec:Methodology}

The proposed approach is shown in Algorithm~\ref{alg:balance_data}. Let \( D = \{(x_i, y_i)\}_{i=1}^{N} \) represent the original imbalanced training dataset, where \(x_i\) denotes training samples and \( y_i \in \{0,1\} \) represent the class labels. Our objective is to obtain a balanced training dataset denoted as $ D_{balanced}= \{(x_i', y_i')\}_{i=1}^{N'}$  with \( N' \geq N \), where each class is equally represented. 
The process begins with SMOTE~\cite{chawla2002smote} to generate synthetic samples by interpolating between a minority sample \(x_i\) and its \(k\)-nearest neighbors.

Then,
to enhance the quality of the balanced data (previous minority class), ENN~\cite{wilson1972asymptotic} is employed to identify and eliminate noisy samples. Specifically, for each $x_j \in D_{\text{balanced}}$, the three nearest neighbors are identified and checked whether $x_j$ is misclassified by these neighbors. If $x_j$ is misclassified, it is eliminated from $ D_{\text{balanced}}$. 
Once class-level balance is achieved, the next step involves addressing any remaining in-between class imbalances and reducing the impact of noisy samples. To accomplish this, we apply Mixup at an instance level by creating new instances from pairs of samples $x_i$ and $x_j$ selected from the same classes. A random value \(\lambda\) is sampled from a Beta distribution, and the mixed sample $\tilde{x} = \lambda x_i + (1 - \lambda) x_j$ is generated to obtain a balanced data at both levels.

\subsection{Workflow}
The entire technique outlined in this study to classify breast cancer is displayed in Fig.~\ref{methodology}. The steps followed by the proposed process are discussed below:
\begin{itemize}
    \item \textbf{Processing and Segmenting:} 
    The median filter is used to reduce noise. The background is suppressed to clean up images. The background is not uniform and contains machine-generated labels like CC or MLO. In order to get rid of these artifacts before segmentation, thresholding is used. Using the strategy outlined in~\cite{10.1007/978-3-662-44303-3_14}, the image is then separated into three overlapping segments. The actions are depicted in Fig.~\ref{image processing}.
    \item \textbf{Extraction and Preparation:}
    In this work, Haralick's Texture Features~\cite{haralick1973textural} are extracted for whole and segmented images. These features are selected based on the hypothesis that the normal images are different in texture as compared to the abnormal ones. Thirteen Haralick features are computed using four orientations of the Gray-Level Co-Occurrence (GLCM) matrix corresponding to two diagonal and two adjacent neighbors. As a result, 52 features per segment/image are generated. The dataset is then split into the train, test, and holdout sets. Afterward, the proposed STEM augmentation approach is applied to the train set to balance both classes.  
    \item \textbf{Analysis:}
     Eight ML classifiers are trained. These include  Random Forest, Linear Discriminant Analysis, Quadratic Discriminant Analysis, Lightgbm, Xgboost, Adaboost, K Nearest Neighbour (KNN), and Extra Tree. The top three best classifiers based on AUC are selected and combined through majority voting to create the final predictions on the holdout dataset.
    
\end{itemize}

\begin{table}[htbp]
\caption{The performance analysis of the proposed approach in comparison to the standard augmentation procedure on the evaluated datasets. CL denotes the ensemble of the three best-performing classifiers; these are described in Section~\ref{sec:results}.
}
\label{result}
\setlength{\tabcolsep}{3.8pt}
\begin{tabular}{@{}cccccccc@{}}
\toprule
\textbf{Dataset}                                    & 

\textbf{Approach} & \textbf{Acc} & \textbf{AUC}  & \textbf{Rec} & \textbf{Pre} & \textbf{F1} & \textbf{CL} \\ \hline
\multicolumn{1}{c}{\multirow{9}{*}{$S_{cc}$}}     & ADASYN            & 0.87         & 0.76          & 0.69         & 0.76         & 0.72        & $L_dQE$     \\
\multicolumn{1}{c}{}                              & BSMOTE            & 0.87         & 0.73          & 0.62         & 0.74         & 0.65        & $L_dQE$     \\
\multicolumn{1}{c}{}                              & SMOTE-EEN           & 0.95         & 0.93          & 0.93         & 0.79         & 0.84        & $L_dQE$     \\
\multicolumn{1}{c}{}                              & SMOTE             & 0.88         & 0.77          & 0.78         & 0.64         & 0.68        & $L_dQE$     \\
\multicolumn{1}{c}{}                              & SMOTENC           & 0.88         & 0.82          & 0.82         & 0.66         & 0.70        & $L_dQE$     \\
\multicolumn{1}{c}{}                              & SMOTE TOMEK       & 0.86         & 0.77          & 0.63         & 0.77         & 0.66        & $L_dQE$     \\
\multicolumn{1}{c}{}                              & SVMSMOTE          & 0.87         & 0.73          & 0.74         & 0.62         & 0.65        & $L_dQE$     \\
\multicolumn{1}{c}{}                              & MIXUP             & 0.88         & 0.90          & 0.68         & 0.90         & 0.73        & $L_dQE$     \\
\multicolumn{1}{c}{}                              & STEM              & 0.94         & \textbf{0.96} & 0.77         & 0.97         & 0.84        & $L_dQE$     \\ \hline

\multicolumn{1}{c}{\multirow{9}{*}{$S_{mlo}$}}    & ADASYN            & 0.95         & 0.80          & 0.81         & 0.78         & 0.79        & $EL_iR$     \\
\multicolumn{1}{c}{}                              & BSMOTE            & 0.85         & 0.80          & 0.81         & 0.80         & 0.81        & $EL_iR$     \\
\multicolumn{1}{c}{}                              & SMOTE-EEN           & 0.85         & 0.80          & 0.86         & 0.93         & 0.88        & $L_dQE$     \\
\multicolumn{1}{c}{}                              & SMOTE             & 0.94         & 0.82          & 0.83         & 0.77         & 0.80        & $EL_iR$     \\
\multicolumn{1}{c}{}                              & SMOTENC           & 0.94         & 0.78          & 0.79         & 0.76         & 0.77        & $EL_iX$     \\
\multicolumn{1}{c}{}                              & SMOTE TOMEK       & 0.93         & 0.82          & 0.82         & 0.73         & 0.77        & $EL_iX$     \\
\multicolumn{1}{c}{}                              & SVMSMOTE          & 0.94         & 0.81          & 0.81         & 0.79         & 0.80        & $EL_iR$     \\
\multicolumn{1}{c}{}                              & MIXUP             & 0.88         & 0.81          & 0.82         & 0.66         & 0.70        & $L_dQE$     \\
\multicolumn{1}{c}{}                              & STEM              & 0.89         & \textbf{0.84} & 0.66         & 0.84         & 0.71        & $L_dQE$     \\ \hline
\multicolumn{1}{c}{\multirow{9}{*}{$S_{cc+mlo}$}} & ADASYN            & 0.87         & 0.75          & 0.75         & 0.63         & 0.67        & $EL_iR$     \\
\multicolumn{1}{c}{}                              & BSMOTE            & 0.89         & 0.68          & 0.68         & 0.63         & 0.65        & $EL_iR$     \\
\multicolumn{1}{c}{}                              & SMOTE-EEN           & 0.85         & 0.77          & 0.78         & 0.62         & 0.69        & $EL_iR$     \\
\multicolumn{1}{c}{}                              & SMOTE             & 0.88         & 0.75          & 0.76         & 0.64         & 0.67        & $EL_iR$     \\
\multicolumn{1}{c}{}                              & SMOTENC           & 0.90         & 0.370          & 0.71         & 0.66         & 0.68        & $EL_iX$     \\
\multicolumn{1}{c}{}                              & SMOTE TOMEK       & 0.87         & 0.76          & 0.77         & 0.63         & 0.69        & $EL_iR$     \\
\multicolumn{1}{c}{}                              & SVMSMOTE          & 0.89         & 0.62          & 0.62         & 0.60         & 0.61        & $EL_iR$     \\
\multicolumn{1}{c}{}                              & MIXUP             & 0.87         & 0.76          & 0.77         & 0.63         & 0.69        & $EL_iR$     \\
\multicolumn{1}{c}{}                              & STEM              & 0.83         & \textbf{0.87} & 0.87         & 0.63         & 0.73        & $L_dQE$     \\ \hline
\multicolumn{1}{c}{\multirow{9}{*}{$F_{cc+mlo}$}} & ADASYN            & 0.86         & 0.78          & 0.79         & 0.63         & 0.66        & EQR         \\
\multicolumn{1}{c}{}                              & BSMOTE            & 0.86         & 0.84          & 0.75         & 0.65         & 0.65        & $EL_iR$     \\
\multicolumn{1}{c}{}                              & SMOTE-EEN           & 0.90         & 0.72          & 0.72         & 0.65         & 0.68        & ERX         \\
\multicolumn{1}{c}{}                              & SMOTE             & 0.88         & 0.81          & 0.82         & 0.66         & 0.74        & EQR         \\
\multicolumn{1}{c}{}                              & SMOTENC           & 0.89         & 0.82          & 082          & 0.67         & 0.74        & $EL_iQ$     \\
\multicolumn{1}{c}{}                              & SMOTE TOMEK       & 0.89         & 0.82          & 0.83         & 0.64         & 0.74        & $EL_iR$     \\
\multicolumn{1}{c}{}                              & SVMSMOTE          & 0.89         & 0.82          & 0.82         & 0.66         & 0.74        & EQR         \\
\multicolumn{1}{c}{}                              & MIXUP             & 0.87         & 0.81          & 0.81         & 0.64         & 0.71        & $L_iQL_d$   \\
\multicolumn{1}{c}{}                              & STEM              & 0.84         & \textbf{0.85} & 0.85         & 0.72         & 0.66        & $L_dQE$     \\ \hline
\multicolumn{1}{c}{\multirow{9}{*}{WBC}}          & ADASYN            & 0.95         & 0.94          & 0.94         & 0.95         & 0.96        & $L_dQE$     \\
\multicolumn{1}{c}{}                              & BSMOTE            & 0.95         & 0.94          & 0.94         & 0.95         & 0.95        & $L_dQE$     \\
\multicolumn{1}{c}{}                              & SMOTE-EEN           & 0.95         & 0.94          & 0.94         & 0.95         & 0.95        & $EKL_i$     \\
\multicolumn{1}{c}{}                              & SMOTE             & 0.95         & 0.94          & 0.95         & 0.94         & 0.95        & $L_dQE$     \\
\multicolumn{1}{c}{}                              & SMOTENC           & 0.96         & 0.95          & 0.97         & 0.95         & 0.96        & $L_dQE$     \\
\multicolumn{1}{c}{}                              & SMOTE TOMEK       & 0.95         & 0.94          & 0.94         & 0.95         & 0.95        & $L_dQE$     \\
\multicolumn{1}{c}{}                              & SVMSMOTE          & 0.94         & 0.94          & 0.95         & 0.94         & 0.94        & $L_dQE$     \\
\multicolumn{1}{c}{}                              & MIXUP             & 0.95         & 0.94          & 0.94         & 0.95         & 0.95        & $L_dEL_i$   \\
\multicolumn{1}{c}{}                              & STEM              & 0.98         & \textbf{0.99} & 0.99         & 0.98         & 0.98        & $AKL_r$     \\ \hline
\end{tabular}
\end{table}

\section {Experimental details}
\label{sec:Experimental details}
The DDSM dataset consists of images. We preprocessed this to extract 13 Haralick Features derived from the GLCM matrix, for each of four orientations, giving a total of 52 features. The WBC dataset consists 30 feature vector samples and didn't need to be preprocessed. The experimentation is conducted on Google Colab. To assess the effectiveness of various classifiers, we employ Pycaret~\cite{PyCaret}, a fundamental ML algorithm that facilitates comparison among different classifiers.

For the experimental process, initially, data points are collected and preprocessed to eliminate noise. Subsequently, the dataset is divided into training and testing holdout groups in an 80/10/10 ratio. To address class imbalance issues, various oversampling techniques are applied. It should be noted that all DDSM setups have significant class imbalances, with class ratios ranging from 6:94, as illustrated in Table~\ref{Table:dataset}. The WBC dataset also exhibits class divisions of 37\% and 63\% for positive and negative classes, respectively. The augmented training data is then used to train models available in the Pycaret library. Based on the Area Under the Curve (AUC) metric, we select the top three models and ensemble them using a majority voting approach. Final predictions are made on the holdout dataset, which serves as previously unseen data for the model, having not been used in training. We explore six different oversampling techniques, including SMOTE ~\cite{chawla2002smote}, Borderline SMOTE ~\cite{han2005borderline}, SMOTENC~\cite{chawla2002smote}, SVMSMOTE~\cite{nguyen2011borderline}, ADASYN~\cite{he2008adasyn}. Furthermore, we compare two hybrid methods of SMOTE-EEN ~\cite{wilson1972asymptotic} and SMOTE-Tomek~\cite{zeng2016effective}, against the proposed STEM algorithm. The details of augmented samples generated by different augmentation methods for each dataset are present in Fig.~\ref{sample}. Our proposed approach produces a balanced number of samples for each class. In comparison to other methods, STEM has the ability to increase the number of data samples more broadly.

\begin{figure*}[t]
    \centering
    \begin{minipage}{0.24\textwidth}
        \centering
        \includegraphics[width=5.5cm, height=5cm]{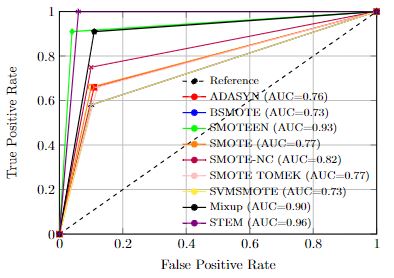}
        \centering (a)
    \end{minipage}%
    \hspace{0.09\textwidth} 
    \begin{minipage}{0.24\textwidth}
        \centering
        \includegraphics[width=5.5cm, height=5cm]{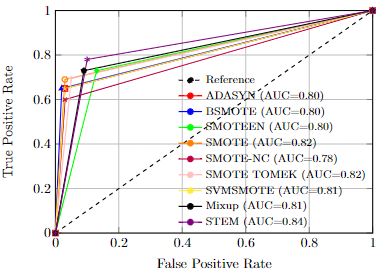}
        \centering (b)
    \end{minipage}%
    \hspace{0.09\textwidth} 
    \begin{minipage}{0.24\textwidth}
        \centering
        \includegraphics[width=5.5cm, height=5cm]{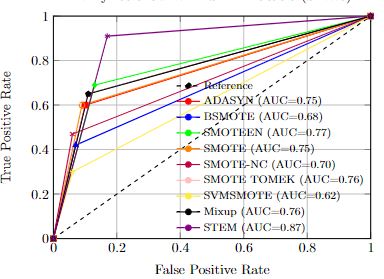}
        \centering (c)
    \end{minipage}
    \hspace{0.09\textwidth} 
    \begin{minipage}{0.22\textwidth}
        \centering
        \includegraphics[width=5.5cm, height=5cm]{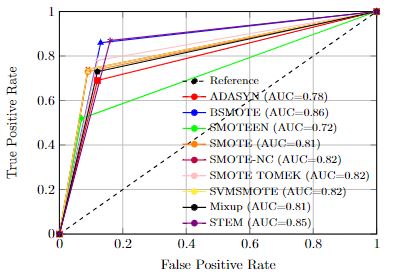}
        \centering (d)
    \end{minipage}%
    \hspace{0.09\textwidth} 
    \begin{minipage}{0.22\textwidth}
        \centering
        \includegraphics[width=5.5cm, height=5cm]{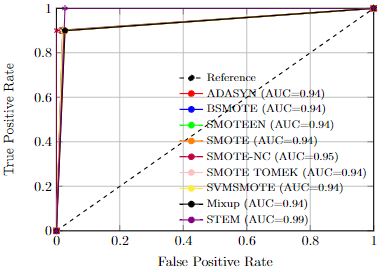} 
        \centering (e)
    \end{minipage}

    \caption{Performance analysis of Area under Receiver Operating Characteristic Curve}
    \label{ROCcurve}
\end{figure*}

\section {Results and Discussion}
\label{sec:results}
 In the presence of a high imbalance problem, accuracy is not considered an appropriate evaluation metric. Instead, Area Under the Curve (AUC), Precision, Recall, and F1-Score are utilized to assess the outcomes of each classification mode~\cite{tohka2021evaluation}. The evaluation metrics are calculated using the following equations (~\ref{eq1}-~\ref{eq4}), where T\textsubscript{Pos} represents true positive cases, T\textsubscript{Neg} denotes samples correctly diagnosed as negative, F\textsubscript{Pos} indicates instances incorrectly classified as positive, and F\textsubscript{Neg} represents positive data points misclassified as negative.
\begin{small}
\begin{equation}\label{eq1}
A_{cc} = \frac{T_{Pos} + T_{Neg}}{T_{Pos} + F_{Neg} + F_{Pos} + T_{Neg}}
\end{equation}

\begin{equation}\label{eq2}
Precision = \frac{T_{Pos}}{T_{Pos} + F_{Pos}}
\end{equation}

\begin{equation}\label{eq3}
Recall = \frac{T_{Pos}}{T_{Pos} + F_{Neg}}
\end{equation}

\begin{equation}\label{eq4}
F1\text{-}Score = 2 \times \frac{Precision \times Recall}{Precision + Recall}
\end{equation}
\end{small}

The ensemble classifiers are denoted by their respective initials: $L_d$ for Linear Discriminant Analysis, $Q$ for Quadratic Discriminant Analysis, $E$ for ExtraTree, $R$ for Random Forest, $L_i$ for Lightgbm, $K$ for KNN, $A$ for Adaboost, and $X$ for Xgboost. Table~\ref{result} presents the results for the $S_{cc}$ setup, where the ensemble of Linear Discriminant Analysis, Quadratic Discriminant Analysis, and Extra Tree Classifier ($L_iQE$) achieved the highest AUC of 0.96 when samples were augmented using our proposed STEM technique. On the other hand, the AUC was lowest at 0.73 for the ensemble created by applying the borderline SMOTE and SVMSMOTE methods. In the case of $S_{mlo}$, STEM outperforms other methods with an AUC of 0.84 when using the ensemble of ($L_iQE$) classifiers. Conversely, the SVMSMOTE sampling approach struggled with the lowest AUC of 0.62 when using the ensemble of ExtraTree, Lightgbm, and Random Forest classifiers ($EL_iR$).

In the case of the configuration $S_{cc+mlo}$, which contains both CC and MLO views of segments as shown in Table~\ref{result}, the proposed STEM technique yielded the maximum AUC of 0.87. The classifiers that performed best were the ones based on ($L_iQE$). Conversely, the SVMSMOTE method of oversampling, in combination with the ensemble of ($EL_iR$) classifiers, achieved the lowest AUC of 0.62. For the final DDSM setup $F_{cc+mlo}$, STEM again achieved the highest AUC of 0.85 as presented in Table~\ref{result}. On the other hand, SMOTE-ENN showed poor performance with an AUC of 0.72. The classifiers for STEM and SMOTE-ENN were based on ($L_iQE$) and ExtraTree, Random Forest, and Xgboost ($ERX$), respectively. Regarding the WBC dataset in Table~\ref{result}, an AUC of 0.99 was obtained by using the STEM-generated samples. The ensemble classifiers used were Adaboost, KNN, and Logistic Regression ($AKL_r$). With the exception of SMOTE-NC, all other approaches achieved an AUC of 0.94. The results reported are obtained solely from the augmented datasets. This is because the non-augmented datasets contain a high-class imbalance ratio, which resulted in poor outputs. The model is biased towards the negative class and is unable to differentiate between the two.

AUC is considered a significant metric for assessing the performance of a binary classifier. This metric is derived by analyzing the Receiver Operating Characteristic (ROC) curve, which plots the True Positive Rate (TPR) against the False Positive Rate (FPR). The performance analysis of each dataset is displayed in Fig.~\ref{ROCcurve}. It is clear that our proposed method surpasses all other augmentation approaches.


\section{Conclusions and Future Work}
\label{sec:conclude}

This study presents a novel data balancing technique, STEM, specifically designed to address class imbalance issues by harnessing the collective power of an ensemble of diverse ML classifiers alongside Mixup  as a generic vicinal distribution. To assess the algorithm's performance comprehensively, eight oversampling and hybrid methods, including SMOTE, SMOTE-NC, SMOTE TOMEK, SMOTE-EEN, ADYSAN, Borderline SMOTE, and SVMSMOTE are selected. Moreover, we employ two publicly available datasets: DDSM with four different setups and WBC. Results show that the proposed approach outperforms all other setups, achieving an outstanding AUC of 0.96 and 0.99 on the DDSM ($S_{cc}$) and WBC datasets, respectively. Notably, the ensemble of the top three classifiers—Linear Discriminant Analysis, Quadratic Discriminant Analysis, and Extra Tree yielded the best results for the $S_{cc}$ setup, while Adaboost, KNN, and Logistic Regression performed admirably for the WBC dataset.
As a future work,  extracting additional image features such as wavelet transform and local binary pattern to enrich the feature vector and generate more diverse samples would enhance the overall performance. Furthermore, exploring the combination of different datasets is valuable to assess the robustness of the proposed approach across variable image data.

\section*{\uppercase{Acknowledgements}}
This study was funded by the Science Foundation Ireland (SFI) Centre for Research Training in Artiﬁcial Intelligence (CRT-AI) Grant No. 18/CRT/6223 and the Irish Software Engineering Research Centre (Lero) Grant No. 16/IA/4605.

\end{document}